\documentclass{article}
\usepackage{amsmath,epsfig}
\usepackage[preprint]{spconfa4}
\usepackage[bookmarks=false]{hyperref}

\copyrightnotice{978-1-7281-1331-9/20/\$31.00~\copyright 2020 IEEE}

\let\OLDthebibliography\thebibliography
\renewcommand\thebibliography[1]{
  \OLDthebibliography{#1}
  \setlength{\parskip}{0pt}
  \setlength{\itemsep}{0pt plus 0.3ex}
}

\usepackage{graphics,caption}
\usepackage{cuted}
\usepackage{float}
\usepackage{algorithm}
\usepackage{algpseudocode}  
\usepackage{amsmath}  
\usepackage{amssymb}
\usepackage{hyperref}
\usepackage{color}
\usepackage[all]{hypcap}
\begin{document}\sloppy
	
	\def\x{{\mathbf x}}
	\def\L{{\cal L}}
	\def\lyu{\textcolor{red}}
	
	\def\mr{\mathrm}
	\def\mb{\mathbf}
	\def\mbb{\mathbb}
	\def\mc{\mathcal}
	\def\mt{\mathtt}
	\def\diag{\mbox{diag}}
	\def\rank{\mbox{rank}}
	\def\grad{\mbox{\text{grad}}}
	\def\dist{\mbox{dist}}
	\def\sgn{\mbox{sgn}}
	\def\tr{\mbox{tr}}
	\def\etal{{\em et al.\/}\, }
	\def\card{{\mbox{Card}}}
	\def\st{\mbox{s.t. }}
	\def\ie{\textit{i.e.}}
	
	\graphicspath{{figs/}}

\title{Modeling Cross-view Interaction Consistency \\for Paired Egocentric Interaction Recognition}
%
\name{Zhongguo Li$^{\ast}$, Fan Lyu$^{\ast}$, Wei Feng$^{\ast}$, Song Wang$^{\ast\dagger}$}
\address{$^{\ast}$College of Intelligence and Computing, Tianjin University, China\\$^{\dagger} $Department of Computer Science and Engineering, University of South Carolina, USA\\ \{lizhongguo,fanlyu\}@tju.edu.cn, wfeng@ieee.org, songwang@cec.sc.edu}

\maketitle

\begin{abstract}
	With the development of Augmented Reality (AR), egocentric action recognition (EAR) plays an important role in accurately understanding demands from the user.
	However, EAR is designed to help recognize human-machine interaction in single egocentric view, thus difficult to capture interactions between two face-to-face AR users.
	Paired egocentric interaction recognition (PEIR) is the task to collaboratively recognize the interactions between two persons with the videos in their corresponding views.
	Unfortunately, existing PEIR methods always directly use linear decision function to fuse the features extracted from two corresponding egocentric videos, which ignore the 
	consistency of interaction in paired egocentric videos. 
	The consistency of interactions in paired videos, and features extracted from them, are correlated to each other. 
	On top of that, we propose to derive the relevance between two views using bilinear pooling, which captures the consistency of two views in feature-level. 
	Specifically, each neuron in the feature maps from one view connects to the neurons from the other view, which 
	enforces the compact consistency between two views and then all possible paired neurons are used for PEIR.
	To be efficient, we use compact bilinear pooling with Count Sketch to avoid directly computing outer product. 
	Experimental results on the PEV dataset shows the superiority of the proposed methods on the task PEIR.
	\let\thefootnote\relax\footnotetext{This work was supported, in part, by the National Natural Science Foundation of China NSFC-U1803264, NSFC-61672376, and NSFC-61671325.}
\end{abstract}

\section{Introduction}
Due to the advance of Augmented Reality (AR) techniques, wearable AR devices like Microsoft HoloLens allow users to interact with the real world via gestures or voice commands. 
Egocentric action recognition (EAR)~\cite{li2019deep,Choutas2018PoTion,Sudhakaran2018LSTA,ryoo2015pooled} is the task to recognize the action or gesture of users to achieve intelligent human-machine interaction.
In this paper, we study the further problem of Paired Egocentric Interaction Recognition (PEIR) that recognizes the interactions between face-to-face AR users \cite{syahputra2018interaction,karambakhsh2019deep,pan2018and}.
Different from EAR that only considers one egocentric video, PEIR needs to simultaneously consider the paired face-to-face egocentric videos.
Utilizing paired egocentric videos recorded from face-to-face views can obtain more precise recognitions than egocentric videos from single views~\cite{yonetani2016recognizing}. 
Enabling AR systems to understand the interactions between persons can provide more precise assistance in daily life. 
For examples, while one user point at one object and the other one shares his/her attention, AR system could read the interaction and response without explicit commands.

\begin{figure}[t]
	\centering
	\includegraphics[width=.95\linewidth]{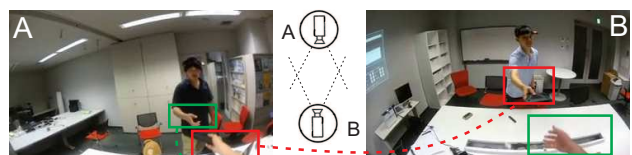}
	\vspace{-10px}
	\captionof{figure}{An illustration of the consistency of interactions between a paired of egocentric views. 
		Here the right hand of person $\text{A}$, labeled by red boxes, are recorded in both the views of $\text{A}$ and $\text{B}$. }
	\vspace{-16px}
	\label{fig_intro}
\end{figure}

Previous works make many efforts on EAR~\cite{li2019deep,Choutas2018PoTion,Sudhakaran2018LSTA,ryoo2015pooled}.
Ryoo et al.~\cite{ryoo2015pooled} tries to capture both entire scene dynamics and salient local motions observed in videos to predict interactions. Li et al.~\cite{li2019deep} 
tries to model the relationship of the camera wearer and the interactor.
From EAR methods, a naive solution to PEIR is directly fusing the features from two views for the sake of one head output.
For instance, one can concatenate the features extracted from paired egocentric videos and directly use the generated feature for linear classification.
Till now, few works try to solve the PEIR problem.
In~\cite{yonetani2016recognizing}, Yonetani et al. adopt a linear decision function to combine two kinds of handcrafted feature for PEIR. 

In our view, the naive feature fusion from two views ignores the interaction consistency in two videos taken from two persons' views respectively.
As shown in Fig.~\ref{fig_intro}, we consider two persons of A and B and their paired views are shown as A and B respectively. 
On the one hand, the head's tilting of $\text{A}$ could be recorded in the view of $\text{B}$, and also leads to the shift of $\text{A}$'s viewpoints, because cameras are mounted over their heads.
On the other hand, there may be interactions occurred in common areas recorded in both views of A and B. 
In the both cases, there are explicit or implicit information consistency of interactions between the views of A and B.

Due to the consistency of interactions in two views, neurons of feature maps corresponding to two views should describe the same information. 
We propose to classify interactions based on the consistent information represented by all possible paired neurons. 
Specifically, we use bilinear pooling, a second order polynomial kernel function~\cite{gao2016compact}, to capture the compact consistent information of interaction in all pairs. 
To avoid directly computing expensive outer product, we propose to use compact bilinear pooling to reduce the computation cost by using Count Sketch.
We first extract features from paired egocentric videos, followed by obtaining sketch mentioned above and transforming them into Fourier domain. Finally we
compute element-wise product of them and transform the result into real domain for linear classification.
Experimental results on the PEV dataset shows that the proposed method outperforms other methods using naive fusions.



%

\section{Related Works}

\textbf{Paired egocentric interaction recognition (PEIR)}, extended from egocentric action recognition (EAR), aims at recognizing the interaction between two face-to-face persons
from both of their views. Along with the successes of deep learning in image-level tasks ~\cite{ISI:000457843607093,ISI:000476809700007}, 
in recent years, EAR makes a great progress by using deep neural networks.
For example 
Li et al.~\cite{li2019deep} models the relationship of the camera wearer and the interactor.
However, EAR assumes there exists only one person with wearable camera.
PEIR was first proposed by Yonetani et al.~\cite{yonetani2016recognizing}, where interactions like subtle motion of head or small hand actions people used in communications
are recognized, by using their paired face-to-face views. 
The video from each view is collected by the camera over the head. 
To recognize interaction, Yonetain et al.~\cite{yonetani2016recognizing} combines two kinds of hand-crafted features -- PoTCD~\cite{poleg2014temporal} for head and iDT~\cite{wang2013action} for body --
for two views respectively with linear decision models.
However, it ignores the consistency of interactions in two views and prefers to rely on only one view for prediction. 
Different from naive fusion like concatenation, we propose to leverage bilinear pooling to model the consistency between the 
paired egocentric videos.

\noindent
\textbf{Bilinear pooling} models~\cite{lin2015bilinear} was proposed for fine-grained image classification. 
Bilinear methods have been used to fuse two kind of features extracted by deep neural networks. 
Given two features $\mb{f}^1, \mb{f}^2 \in \mathbb{R}^{C}$, bilinear methods compute the outer product of them by
\begin{equation}
\label{eq_bilinear}
\mb{f}^\text{1} \otimes \mb{f}^\text{2} = [{\mb{f}^\text{1}}  \times {(\mb{f}^\text{2})}^T ] \in \mathbb{R}^{C^2}, 
\end{equation}
where $\otimes$ means outer product of tensors.  
$[\,\cdot\,]$ operator vectorizes a matrix into a vector which means elements in matrix is sorted by orders in new vector. 
Then the features generated by bilinear methods are directly used for classification. 
However when $C$ is large, the generated features are of dimension $C^2$. 
High-dimensional features make the direct computing of the outer product for linear classification very expensive. 
Gao et al.~\cite{gao2016compact} reduces the dimension of generated feature from $C^2$ into $D$, which is far less than $C^2$.
Following this idea, Fukui et al.~\cite{fukui2016multimodal} fuse the visual features and the text features for visual question answering and visual grounding.

\section{Method}

\subsection{Problem formulation}

Denote the person who starts the interaction as $\text{A}$,  the one who receives the interaction as $\text{B}$, as shown in Fig.~\ref{fig_intro} and
their recorded videos (taken by the wearable cameras mounted over their heads) as $v^\text{A}$ and $v^\text{B}$, respectively.
PEIR tries to learn to recognize the interaction $l^*$ from such paired egocentric videos based on the hypothesis $l^*=h(v^\text{A}, v^\text{B})$. 

The naive method to tackle PEIR is directly fusing the information from two views, such as \cite{yonetani2016recognizing}, and this method can be formulated as 
\begin{align}
\label{model_def}
&\mb{f}^\text{A} = \text{CNN}(v^\text{A}),\\ 
&\mb{f}^\text{B} = \text{CNN}(v^\text{B}), \\
&\mb{g} = \Phi(\mb{f}^\text{A}, \mb{f}^\text{B}), \label{def_phi} \\
&p(l|v^\text{A},v^\text{B}) = \sigma(\mb{W}^\top\mb{g}  + \mb{b}), \label{def_softmax}
\end{align}
where $\mb{f}^\text{A}, \mb{f}^\text{B} \in \mathbb{R}^{C}$ are the flattened feature maps extracted by Convolutional Neural Network (CNN) and $\mb{g} \in \mathbb{R}^{D}$ is the fused feature of $\mb{f}^\text{A}$ and $\mb{f}^\text{B}$, $\sigma$ is the softmax activation function. 
The key step is the design of the fusing function $\Phi$ to effectively combine the information from two views.
Common choices of the fusing function $\Phi$ include

	1) concatenation: $\Phi(\mb{x},\mb{y})=[\mb{x},\mb{y}]$,
	
	2) element-wise summation: $\Phi(\mb{x},\mb{y})=\mb{x} + \mb{y}$, and
	
	3) element-wise product: $\Phi(\mb{x},\mb{y})= \mb{x} \odot \mb{y}$.

These three common fusing methods are easy to implement but do not consider the possible interaction consistency between two views.
In this following, we propose a new method to address this problem.

\subsection{Main Method}
\begin{figure*}
	\centering
	\includegraphics[width=0.9\linewidth]{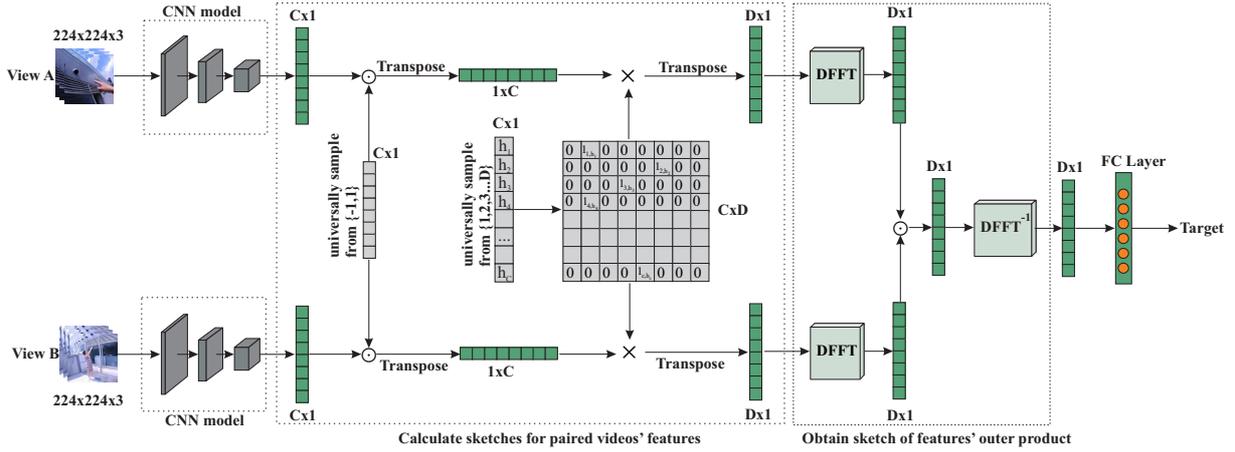}
	\caption{Framework of the proposed method. Given paired egocentric videos from views A and B, we first extract features from both videos, then calculate the sketches of them, finally we obtain the sketch of outer product and transform it into Fourier domain for classification. $\text{DFFT}$ is the discrete fast Fourier transform.}
	\label{fig_bi3d}
\end{figure*}
The consistency of interactions exist in paired egocentric videos as the common visual information of co-viewed objects, i.e.,  $\mb{f}^\text{A}$ and $\mb{f}^\text{B}$ are correlated.
We first enumerate all pairs between $\mb{f}^\text{A}$ and $\mb{f}^\text{B}$ by
\begin{equation}
\label{def_relevance}
\mb{g}_{k} = \phi(\mb{f}^\text{A}_i, \mb{f}^\text{B}_j) ,
\end{equation}
where $\mb{g} \in \mathbb{R}^{C^2}$ represents the consistent information in the form of pairs and $k = i \cdot C + j$ represents the $k$-th pair between $\mb{f}^\text{A}$ and $\mb{f}^\text{B}$.
$\mb{f}_i$ is the $i$-th element in $\mb{f}$ and $\phi$ is element-wise form of $\Phi$. 
By Eq.~\eqref{def_relevance}, we construct element-wise information correlation between view A and view B, which yields fine-grained interaction consistency representation $\mb{g}$.
In Eq.~\eqref{def_relevance}, if the product operator is adopted for $\phi$, $\mb{g}$ is actually the outer product of $\mb{f}^\text{A}$ and $\mb{f}^\text{B}$, and $\Phi$ becomes the bilinear method as described 
in Eq.~\eqref{eq_bilinear}.
The procedure of the compact bilinear method is shown in Alg.~1 in the appendix.
To reduce the computation cost, we use the compact bilinear method~\cite{pham2013fast,gao2016compact,fukui2016multimodal} to shrink the dimension of $\mb{g}$ from $C^2$ to $D$, where $D \ll C^2$, and the full framework is shown in Fig.~\ref{fig_bi3d}. Gao et al.~\cite{gao2016compact} proves that bilinear method is actually a second order polynomial kernel function and uses Count Sketch (called sketch in the remainder of this paper)~\cite{gao2016compact, pagh2013compressed} to reduce the computation cost.

Detailedly, we first calculate the sketch of feature $\mb{f}^\text{A}$ and $\mb{f}^\text{B}$. We define $\mb{h},\mb{s}$ as following:
each element in $\mb{h} \in \mathbb{N}^C_+$ is universal randomly sampled from $\{1,2,3, \cdots, D\}$ and every element in $\mb{s} \in \mathbb{Z}^C$ is universal randomly sampled from $\{-1,1\}$. 
We denote $\mb{H}$ as a $C \times D$ matrix generated according to $\mb{h}$: each element $\mb{H}_{i,j}$ is defined as
\begin{align}
\mb{H}_{i,j} &= \left\lbrace
\begin{array}{rcl}
1,       &{\text{if } j = \mb{h}_i},\\
0,		&\text{otherwise}.\\
\end{array}
\right. 
\end{align}
For the feature $\mb{f}$, we first compute element-wise product of $\mb{f}$ and $\mb{s}$, similar to traditional Random Maclaurin \cite{kar2012random} for approximating the polynomial kernel. 
Then we compute matrix product of the above element-wise product and matrix $\mb{H}$ to get sketch of feature $\mb{f}$, which projects $\mb{f}$ from $\mathbb{R}^C$ into $\mathbb{R}^D$ to provide bounds on the variance of estimates to guarantee the reliability \cite{pham2013fast}:
\begin{align}
\mb{sketch}^\text{A} &= (\mb{s}^\text{A} \odot \mb{f}^\text{A}) \mb{H}^\text{A}, \\
\mb{sketch}^\text{B} &= (\mb{s}^\text{B} \odot \mb{f}^\text{B}) \mb{H}^\text{B}.
\end{align}
After that, we transform sketches into Fourier domain using discrete fast Fourier transform ($\text{DFFT}$) and obtain element-wise product result of two sketches in Fourier domain. It has been proved~\cite{pham2013fast} that the sketch of outer product $\mb{f}^\text{A} \otimes \mb{f}^\text{B}$ in Fourier domain is just the element-wise product of  $\mb{f}^\text{A}$'s sketch $\mb{sketch}^\text{A}$ and  $\mb{f}^\text{B}$'s sketch $\mb{sketch}^\text{B}$ both in Fourier domain, and the computation of outer product could be replaced by computing element-wise product in Fourier domain. Finally we obtain $\mb{g}$ in the original domain by transforming the result in Fourier domain back using inverse DFFT ($\text{DFFT}^{ -1 }$), i.e.,
\begin{align}
\mb{g} &=\text{DFFT}^{ -1 } (\text{DFFT}(\mb{sketch}^\text{A}) \odot \text{DFFT}(\mb{sketch}^\text{B})).
\end{align}
%
%
%

\subsection{Discussion}
Although the concatenation method and element-wise operations can also be written in the form of Eq.~\eqref{model_def}, they cannot describe the consistency of interaction in two views. 
The reasons are discussed as follows.

\noindent
\textbf{Concatenation.}
For concatenation methods, the fusing function $\Phi$ in Eq.~\eqref{def_phi} is to concatenate $\mb{f}^\text{A}$ and $\mb{f}^\text{b}$, and the logits of the PEIR prediction can be written as
\begin{align}
p(l|v^\text{A},v^\text{B}) &\propto \prod_{i=1}^{2C} \exp\left({\mb{W}_i^\top \mb{g}_i}\right),  \label{def_cat_1} \\
&= \prod_{i=1}^{C} \exp\left({\mb{W}_i^\top \mb{f}^\text{A}_i}\right)\prod_{i=C+1}^{2C} \exp\left({\mb{W}_i^\top \mb{f}^\text{B}_j}\right). \label{def_cat_2}
\end{align}
Since the first $C$ elements of $\mb{g} \in \mathbb{R}^{2C}$ generated by the concatenation method are all from feature $\mb{f}^\text{A}$ and the rest are all from $\mb{f}^\text{B}$, we could rewrite the predicted probability of the model defined in Eq.~\eqref{def_softmax} as Eq.~\eqref{def_cat_2}.
Obviously, in Eq.~\eqref{def_cat_2}, the feature $\mb{f}^\text{A}$ is independent of $\mb{f}^\text{B}$ for final prediction. 
Thus concatenation method does not consider the correlation between $\mb{f}^\text{A}$ and $\mb{f}^\text{B}$.

\noindent
\textbf{Element-wise product.}
Element-wise product is similar to Eq.~(\ref{def_relevance}), and the predicted probability could be written as
\begin{align}
&p(l|v^\text{A},v^\text{B}) \propto \prod_{i=1}^{C} \exp\left({\mb{W}_i^\top \mb{f}^\text{A}_i \mb{f}^\text{B}_i}\right),
\end{align}
where only feature elements with the same index is correlated.

\noindent
\textbf{Element-wise summation.}
If we adopt element-wise summation as $\phi$, we could see 
\begin{eqnarray}
\begin{aligned}
p(l|v^\text{A},v^\text{B}) &\propto \prod_{i=1}^{C} \exp\left({\mb{W}_i^\top (\mb{f}^\text{A}_i+\mb{f}^\text{B}_i)}\right) \\
&= \prod_{i=1}^{C} \exp\left({\mb{W}_i^\top \mb{f}^\text{A}_i}\right) \prod_{i=1}^{C} \exp\left({\mb{W}_i^\top \mb{f}^\text{B}_i}\right).\\
\end{aligned}
\end{eqnarray}
Similar to the concatenation-based fusion, in this case features $\mb{f}^\text{A}$ and $\mb{f}^\text{B}$ are independent of each other for final prediction. 

\noindent
\textbf{Ours.}
 We adopt bilinear pooling operation as $\Phi$, which means the product operator is selected as $\phi$ in Eq.~(\eqref{def_relevance}), i.e., 
\begin{eqnarray}
\begin{aligned}
p(l|v^\text{A},v^\text{B}) &\propto \prod_{i=1,j=1}^{C} \exp\left({\mb{W}_i^\top \mb{f}^\text{A}_i \mb{f}^\text{B}_j}\right),
\end{aligned}
\end{eqnarray}
where the logits of PEIR are predicted based on the paired elements between $\mb{f}^\text{A}$ and $\mb{f}^\text{B}$. 
Each element in $\mb{f}^\text{A}$ is correlated with all the elements in $\mb{f}^\text{B}$ for final prediction.

\begin{table}[htbp]
	\centering	
	\caption{Accuracy in validation set with RGB or OpticalFlow.}
	\label{fig_result}
	\resizebox{\linewidth}{!}{
		\begin{tabular}{c c c c | c c c c}
			\hline
			Modality	&View	&Method 	&$D$ &Split1 &Split2 &Split3 &Avg Acc\\
			\hline
			RGB &A	& &	-			&78.50	&77.94	&76.19	&77.54\\
			RGB &B 	& &	-		&73.52	&68.73	&71.75	&71.33\\
			RGB &A+B &Avg &	-		&80.50	&80.55	&76.80	&79.28\\
			RGB &A+B &Svm	& -		&\textbf{84.96}	&78.70	&80.40	&81.35\\
			RGB &A+B &Concat & 2048 	&82.25	&80.89	&79.30	&80.81\\
			RGB &A+B &Ours & 2048	&84.35	&\textbf{83.09}	&\textbf{83.47}	&\textbf{83.63}\\
			
			\hline
			OF &A &	&-			&84.96	&81.35	&84.24	&83.51\\
			OF &B &	&-			&79.74	&78.00	&78.46	&78.73\\
			OF &A+B &Avg &-		&86.01	&82.62	&85.86	&84.83\\
			OF &A+B &Svm &-		&\textbf{89.40}	&83.42	&87.97	&86.93\\
			OF &A+B &Concat	&2048	&87.37	&87.56	&88.21	&87.71\\
			OF &A+B &Ours &2048	&88.10	&\textbf{89.06}	&\textbf{89.12}	&\textbf{88.76}\\
			\hline
			RGB &A+B &Sum  &1024	&82.99	&79.14	&79.09	&80.41\\
			RGB &A+B &Product &1024 &84.35	&79.10	&72.81	&78.75\\
			RGB &A+B &Ours &1024  	&\textbf{86.08}	&\textbf{81.34}	&\textbf{83.78}	&\textbf{83.73}\\
			\hline
			OF &A+B &Sum  &1024	&87.24	&87.26	&87.64	&87.78\\
			OF &A+B &Product &1024	&87.30	&88.44	&88.34	&88.03\\
			OF &A+B &Ours &1024 &\textbf{89.33}	&\textbf{88.81}	&\textbf{88.37}	&\textbf{88.84}\\		
			\hline
		\end{tabular}
	}	
\end{table}


\section{Experiments}

\begin{table}[htbp]
	\centering
	\caption{Mean accuracy of TwoStream nets.}
	\label{fig_result_rgbof}
	\setlength{\tabcolsep}{0.5ex}{
		\begin{tabular}{c c c c | c}	
			\hline
			Modality &View &Method &$D$				&Avg Acc\\
			\hline 
			PoTCD+IDT	&A+B	&Yonetani et  al.\cite{yonetani2016recognizing}  &- &69.2\\
			\hline
			TwoStream &A+B &Svm &-			&87.27\\
			TwoStream &A+B &Concat &2048	&87.72\\
			TwoStream &A+B &Ours &2048			&\textbf{88.91}\\
			\hline
			TwoStream &A+B &Product &1024			&85.17\\
			TwoStream &A+B &Sum &1024	&84.41\\
			TwoStream &A+B &Ours &1024			&\textbf{88.45}\\
			\hline
		\end{tabular}
	}
\end{table}

\begin{figure}[htbp]
	\centering
	\includegraphics[width=\linewidth]{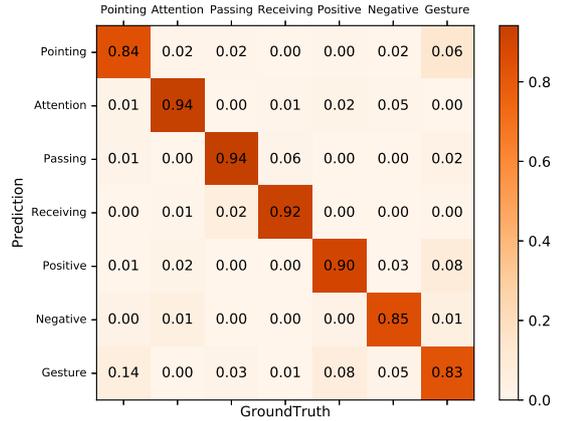}
	\vspace{-25px}
	\caption{Confusion matrix of our method, averaged over the three validation sets.}
	\label{fig_cm}
	\vspace{-10px}
\end{figure}

\begin{figure*}[h]
	\centering
	\includegraphics[width=1.0\linewidth]{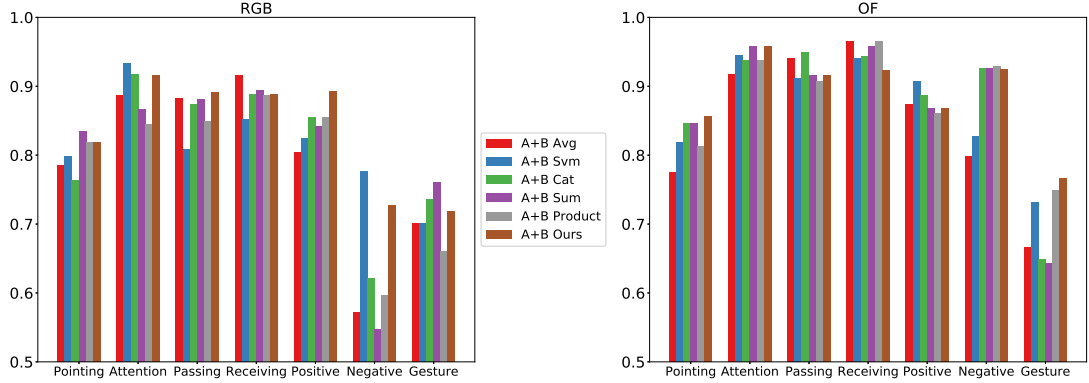}
	\captionof{figure}{Accuracy for each category while input is RGB (left) or OpticalFlow (right). `A+B' means both views are used.}
	\label{acc_each_class}
	\vspace{-0px}
\end{figure*}

\begin{figure*}[h]
	\centering
	\includegraphics[width=1.0\linewidth]{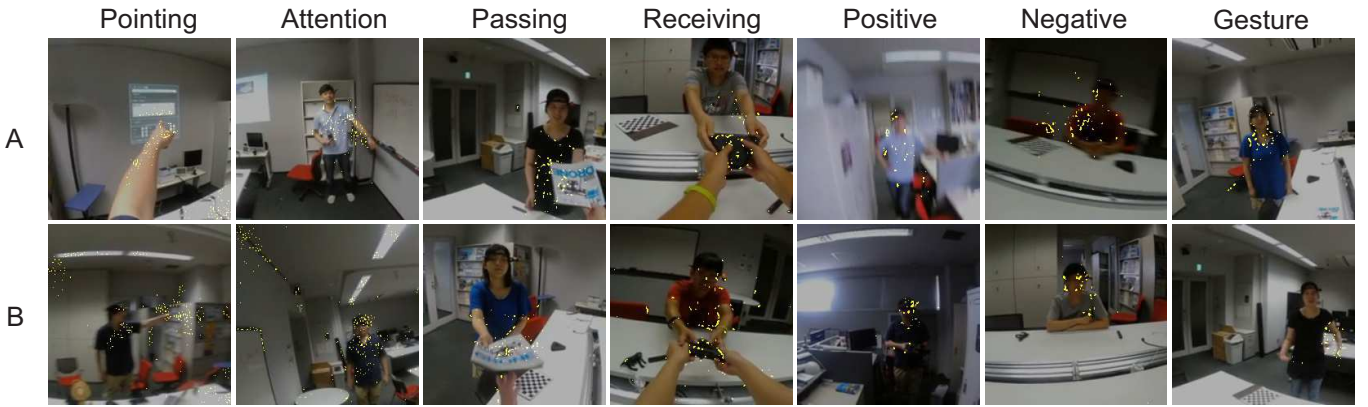}
	\captionof{figure}{Top 10 percent integrated gradients of our model in validation sets. The yellow dots show the model's focuses.}
	\label{fig_ig}
\end{figure*}

\subsection{Experiments details}
\textbf{Dataset and evaluation metric.}
PEV dataset is proposed in~\cite{yonetani2016recognizing}, which contains 1,226 pairs of videos in 7 categories. Each paired video is collected by the cameras mounted to the heads of two persons standing face to face. 
The 7 categories of interactions present in PEV dataset\cite{yonetani2016recognizing} are:
${Pointing}$,
${Attention}$,
${Positive}$,
${Negative}$,
${Passing}$,
${Receiving}$,
${Gesture}$.
We adopt the three-fold cross validation for evaluation. 
Results in all three validation splits are reported in Table ~\ref{fig_result}.

\noindent
\textbf{Model settings.}
We adopt I3D \cite{carreira2017quo} as the backbone, which is pretrained on Kinetic-400. $C$, the dimension of $\mb{f}^\text{A}$ and $\mb{f}^\text{B}$, is set to 1,024 as used in backbone. For fair comparison , $D$, the dimension of $\mb{g}$, is set to 2,048 as in the concatenation method, or 1,024 as in the element-wise product or summation methods. Cross entropy loss is chosen as the loss function. Standard SGD is used to train neural networks in this paper, and the learning rate is set to 0.1. 

\noindent
\textbf{Normalization.}
$\mb{f}^\text{A}$ and $\mb{f}^\text{B}$ shall be normalized to be in the form of one-hot encoding, since they are batch normalized~\cite{ioffe2015batch} 
in the backbone and fed into ReLU~\cite{glorot2010deep} 
layer. We only scale them by a constant factor.

\noindent
\textbf{Data Processing.}
We sample one frame for every three frames. Sampled frames are randomly cropped and scaled to $224\times 224$.
Horizontal flipping is then used for data augmentation.
The Length of clip is set to 32, and the last frame will be repeated if clip's length is less than 32. 

\subsection{Results}

We explore different fusion methods under RGB or OpticalFlow modality. We choose feature-fusion methods including concatenation (Concat), element-wise product (Product) and element summation (Sum) as baselines. We also report results of the methods using average (Avg) or SVM for late-fusion, i.e., training classifier for each view first and then fusing scores as the final result. Results are shown in Table \ref{fig_result}.

\noindent
\textbf{Comparisons.}
All the methods using two views together outperform the models that only use one view. 
We can see that, SVM is always a better choice for fusing decision scores. 
Element-wise product achieves a score higher than element-summation in OpticalFlow modality, but lower in RGB modality. 
While under RGB modality, the end-to-end concatenation method does not outperform the score fusion method, e.g., SVM,
the result is opposite when the input becomes OpticalFlow.
Under OpticalFlow modality, element-wise product is surprisingly better than concatenation or element-wise summation, with half parameters in the last fully connected layer 
compared with the concatenation-based method. 
Our method performs best in both RGB and OpticalFlow modalities -- it outperforms the second best method by 2.28\% 
in RGB modality, and by 1.05\% in OpticalFlow modality, with $D$ set as 2,048. 
When $D$ is set to 1,024, our proposed method outperforms the second best by 3.32\% in RGB modality, and by only 0.81\%
(over element-wise product) in OpticalFlow modality. 
A possible explanation is that only motion information is recorded in OpticalFlow modality and the consistency involving appearance cannot be built in OpticalFlow modality.
When following the two-stream net \cite{simonyan2014two} of using both RGB and OpticalFlow modalities as input, our method still achieves highest score than other fusing methods,
as shown in Table.~\ref{fig_result_rgbof}.

\noindent
\textbf{Confusion matrix and category-grained accuracy.}
Confusion matrix of our method is shown in Fig.~\ref{fig_cm} and Fig.~\ref{acc_each_class}. The proposed method performs best or pretty close to the best. Other methods is unstable compared with ours. 
We believe this is because our proposed method makes final prediction only if evidence is found in both paired egocentric videos.
For RGB modality, $Negative$ action is the most difficult to be recognized, where the score-fusion SVM method performs the best, followed by our method as the second. 
For OpticalFlow modality, the most difficult action is $Gesture$ and the proposed method achieves the highest score.

\subsection{Visualization}
We use Integrated Gradients~\cite{sundararajan2017axiomatic} to visualize the model, as shown in Fig.~\ref{fig_ig}. The yellow dots in frames represent the focus of the model. In $Pointing$ category, person A's hand is focused, and other dots in the edges of frame represent the model focus on these pixels to catch shift of attentions, and similar results can be seen in $Attention$ category. The visualization for $Passing$ and $Receiving$ actions shows that the model successfully focuses on the hand and object. For $Positive$ and $Negative$ actions, the focus is changed into A and B's upper body. The interesting result occurs for $Gesture$ action, the model focuses on B's head, possibly because B responds to A with a subtle shaking of head. 

\section{Conclusion}
In this paper, we proposed to build the relevance between paired egocentric videos for interaction recognition. We observed that the consistency of interactions exists in paired videos and features extracted from them are correlated to each other. We proposed to use bilinear pooling to capture all possible consistent information represented in pairs of elements between the features from two views. Moreover we used the compact bilinear pooling with Count Sketch to reduce the algorithm computation complexity. Experiments showed that our method achieves the state-of-the-art performance on the PEV dataset.



\bibliographystyle{IEEEbib}
\bibliography{mypaper}

\end{document}